\documentclass[11pt]{article}

\usepackage[preprint]{acl}
\usepackage{amssymb}
\usepackage{times}
\usepackage{latexsym}
\usepackage{colortbl}
\usepackage{graphicx}
\usepackage{pifont}

\usepackage[T1]{fontenc}

\usepackage[utf8]{inputenc}

\usepackage{microtype}

\usepackage{inconsolata}

\usepackage{graphicx}

\usepackage{amsmath} 
\usepackage{bm}        
\usepackage{booktabs}
\usepackage{multirow}
\usepackage{enumitem}
\usepackage{xcolor}
\usepackage{tcolorbox}
\usepackage{xspace}
\usepackage{eso-pic}   
\newcommand{\methodname}{\textsc{RSTG}\xspace}

\AddToShipoutPictureBG{%
  \ifnum\value{page}=1
    \AtPageUpperLeft{%
      \put(\LenToUnit{2.5cm},\LenToUnit{-2.5cm}){%
        \includegraphics[height=0.8cm]{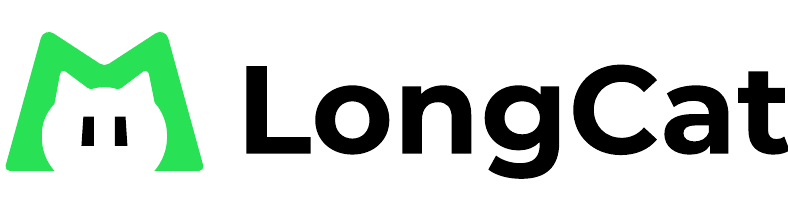}%
      }%
    }%
  \fi
}

\title{Distill Where You Fail: Recovering Learning Signals of Negative RL-Groups from Adaptive Teacher Guidance}

\author{First Author \\
  Affiliation / Address line 1 \\
  Affiliation / Address line 2 \\
  Affiliation / Address line 3 \\
  \texttt{email@domain} \\\And
  Second Author \\
  Affiliation / Address line 1 \\
  Affiliation / Address line 2 \\
  Affiliation / Address line 3 \\
  \texttt{email@domain} \\}

\author{
 \textbf{Zhuowen Han\textsuperscript{1}},
 \textbf{Jinwei Xiao\textsuperscript{2}},
 \textbf{Zhengxi Lu\textsuperscript{2}},
 \textbf{Renren Jin\textsuperscript{1}},
 \textbf{Zhiyuan Yao\textsuperscript{2}},
 \textbf{Yuxin Liu\textsuperscript{2}},
\\
 \textbf{Hongyan Hao\textsuperscript{2}},
 \textbf{Yueqing Sun\textsuperscript{2}},
 \textbf{Yu Yang\textsuperscript{2}},
 \textbf{Qi Gu\textsuperscript{2}},
 \textbf{Xunliang Cai\textsuperscript{2}},
 \textbf{Deyi Xiong \textsuperscript{1,$\dagger$}}
\\
\\
 \textsuperscript{1}TJUNLP Lab, School of Computer Science and Technology, Tianjin University,
 \\
 \textsuperscript{2}Meituan Longcat Team
\\
 \texttt{\small \{zwhan, dyxiong\}@tju.edu.cn\qquad guqi03@meituan.com} 
}

\begin{document}
\maketitle

\begingroup
\renewcommand{\thefootnote}{}

\footnotetext{$\dagger$ Corresponding author}
\endgroup

\begin{abstract}
Reinforcement learning with verifiable rewards (RLVR) has become a standard paradigm for post-training large language models (LLMs). While Group Relative Policy Optimization (GRPO) is widely adopted, it suffers from sparse reward signals and loses gradients entirely when all responses within a group receive identical rewards. On-policy distillation (OPD) offers a natural remedy by providing dense, token-level supervision from a teacher model. However, naively combining GRPO with OPD leads to degraded performance, due to three underlying causes: not all samples benefit from distillation; fitting too quickly to the teacher undermines the exploratory capacity of RL; and OPD's advantages are asymmetric, suppressing most tokens. To address these challenges, we propose \textbf{RSTG} (\textbf{R}ecovering Learning \textbf{S}ignals via Adaptive \textbf{T}eacher \textbf{G}uidance), which applies distillation selectively and precisely where it matters most. At the sample level, OPD is restricted to negative zero-variance prompts with each sample weighted by the teacher's confidence score. At the token level, distillation targets only tokens with high student entropy or large teacher-student divergence. We further augment training with SFT on correct trajectories generated by the teacher model, injecting positive gradient signals where RL yields none. Experiments demonstrate that RSTG substantially outperforms naive GRPO+OPD by +4.02\% on math and +3.05\% on code.
\end{abstract}

\section{Introduction}

Post-training large language models via reinforcement learning with verifiable rewards (RLVR) has emerged as a standard approach for improving reasoning capabilities~\citep{DBLP:journals/corr/abs-2505-09388,DBLP:journals/corr/abs-2501-12948,DBLP:journals/corr/abs-2507-06261}. Among RLVR methods, Group Relative Policy Optimization (GRPO;~\citealp{DBLP:journals/corr/abs-2402-03300}) is widely adopted for its simplicity and stability. GRPO normalizes outcome rewards across a group of rollouts to estimate a scalar advantage applied uniformly to every token, resulting in sparse rewards~\citep{DBLP:journals/corr/abs-2601-07408} and vanishing gradients when all rollouts are correct or incorrect; we refer to such prompts as \textit{positive} and \textit{negative zero-variance prompts}, respectively~\citep{DBLP:journals/corr/abs-2506-02177,DBLP:journals/corr/abs-2510-08696}.

\begin{figure}[t]
  \centering
  \includegraphics[width=\columnwidth]{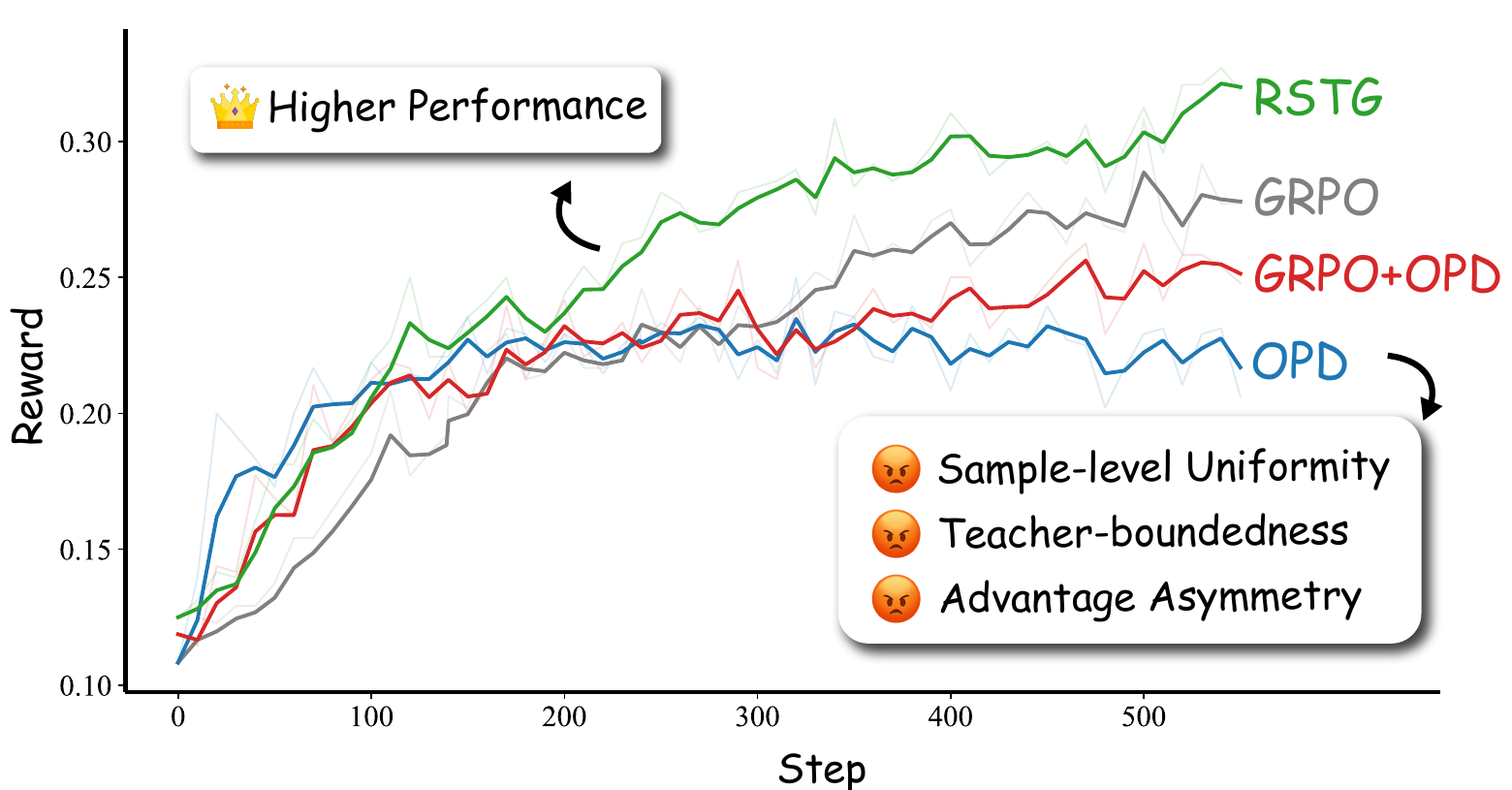}
  \caption{\textbf{Training Dynamics on MATH.} Naively GRPO + OPD proves ineffective, while ours \methodname{} achieves higher performance.}
  \label{fig:motivation}
\end{figure}

Recent work has shifted toward on-policy distillation (OPD;~\citealp{DBLP:conf/iclr/AgarwalVZSGGB24,DBLP:conf/iclr/Gu0WH24}), which has the student generate its own rollouts and leverages the teacher's per-token log-probabilities as a dense reward signal, naturally addressing the limitations of GRPO. A straightforward approach is to complement GRPO with OPD by summing their losses~\citep{DBLP:conf/iclr/AgarwalVZSGGB24}. However, as shown in Figure~\ref{fig:motivation}, this naive combination underperforms standard GRPO, and even with an annealing schedule on the OPD coefficient~\citep{DBLP:journals/corr/abs-2506-02208}, the results remain unsatisfactory.

We identify three key reasons why this naive combination fails: \textbf{(1) Sample-level Uniformity:} Not all samples benefit from OPD; the quality of the teacher's guidance depends on its proficiency on the given sample. \textbf{(2) Teacher-boundedness:} OPD converges quickly with performance capped at the teacher's level. Premature convergence to the teacher severely undermines RL's exploratory capacity. \textbf{(3) Advantage Asymmetry:} Student-generated tokens are typically assigned low probability by the teacher, causing most token-level advantages to be negative~\citep{fu2026revisiting,jia2026asymmetric,DBLP:journals/corr/abs-2603-11137,lu2026self}, suppressing the learning signal. Furthermore, OPD is inherently local, conditioning on a potentially erroneous student-generated prefix and thus producing unreliable gradients~\citep{fu2026revisiting}.

To address the aforementioned limitations, we propose \methodname{} (\textbf{R}ecovering Learning \textbf{S}ignals via Adaptive \textbf{T}eacher \textbf{G}uidance).
\textbf{(1) Sample Selection.} Applying OPD exclusively on prompts where the student fails but the teacher succeeds outperforms standard OPD on all samples, using only $3.63\%$ of the full data. We therefore apply OPD solely on \textit{negative zero-variance prompts}~\citep{DBLP:journals/corr/abs-2509-21880,DBLP:journals/corr/abs-2506-02177} and weight the distillation signal by the teacher's degree of mastery over each prompt.
\textbf{(2) Token Selection.} We restrict OPD gradient updates to tokens where the student exhibits high entropy or where the teacher-student divergence is large, slowing convergence and reducing gradient noise.
\textbf{(3) Auxiliary SFT.} On negative zero-variance prompts, we perform SFT on correct trajectories pre-generated by the teacher, injecting positive gradient signals and providing a global perspective to alleviate the locality of OPD.

We validate \methodname{} across the Qwen2.5 and Qwen3 model families on mathematical and code benchmarks. \methodname{} achieves substantial improvements over naive GRPO+OPD ($+4.02\%$ on mathematics, $+3.05\%$ on code), while also mitigating advantage asymmetry, slowing convergence toward the teacher, and preventing abrupt response length inflation.

In summary, our contributions are as follows:
\begin{itemize}[itemsep=0pt, leftmargin=*]
    \item We identify the applicable scope of OPD and demonstrate the impact of teacher proficiency on its effectiveness.
    \item We propose \methodname{}, enabling effective integration of GRPO and OPD.
\end{itemize}

\section{Related Work}

\paragraph{On-Policy Distillation (OPD).}
Knowledge distillation~\citep{DBLP:journals/corr/HintonVD15} transfers capabilities from a teacher to a student model. OPD~\citep{DBLP:conf/iclr/AgarwalVZSGGB24,DBLP:conf/iclr/Gu0WH24,DBLP:journals/corr/abs-2603-11137} samples trajectories from the student and aligns it with the teacher's token-level logit distribution, providing dense on-policy supervision that naturally complements GRPO. Two dominant paradigms exist: GKD-style OPD uses the token-level KL divergence directly as the training loss~\citep{DBLP:conf/iclr/AgarwalVZSGGB24}, while PG-style OPD treats the per-token reverse KL as a dense reward signal and updates the student via policy gradient, with the advantage defined as the negation of the reverse KL~\citep{DBLP:journals/corr/abs-2603-11137,DBLP:journals/corr/abs-2604-03128}. We adopt the latter, which unifies naturally with the RL framework~\citep{DBLP:journals/corr/abs-2602-12125}.

\paragraph{Combining RL and OPD.}
Prior work has explored integrating RL with knowledge distillation to jointly leverage reward signals and teacher supervision. GKD~\citep{DBLP:conf/iclr/AgarwalVZSGGB24} first explores this unification for text summarization, and KDRL~\citep{DBLP:journals/corr/abs-2506-02208} further extends this direction. More recently, a growing body of work has investigated combining RL with on-policy self-distillation (OPSD)~\citep{DBLP:journals/corr/abs-2604-03128,DBLP:journals/corr/abs-2604-02288,lu2026self}. However, this paradigm still faces significant challenges, and existing methods tend to fail across a broader range of models and settings. \methodname{} aims to make this combination more robust and effective.

\section{Preliminaries}
Let $D$ denote the input distribution, and let $\pi_{\boldsymbol{\theta}}$ and $\pi^{*}$ denote the student and teacher policies, respectively. We unify the following three methods under a common reinforcement learning framework.

\subsection{Group Relative Policy Optimization}
The RL objective can be formulated as
\begin{equation}
    \mathcal{J}_{\mathrm{RL}}(\boldsymbol{\theta})=\max _{\boldsymbol{\theta}} \mathbb{E}_{\boldsymbol{x} \sim D, \boldsymbol{y} \sim \pi_{\boldsymbol{\theta}}(\cdot \mid x)} r(\boldsymbol{x}, \boldsymbol{y}),
\end{equation}
where the trajectories $\boldsymbol{y}$ are sampled from the current policy model $\pi_{\boldsymbol{\theta}}$ , $r(\boldsymbol{x}, \boldsymbol{y})$ is the reward function that measures the quality of a response sequence $\boldsymbol{y}=(y_{1},\cdots,y_{T})$ to a query $\boldsymbol{x}$. A common approach to solving this objective is to apply policy gradient, updating the policy parameters using an estimated gradient of the form:
\begin{equation}
\label{eq:grpo_gradient}
    \begin{aligned}
        \nabla_{\boldsymbol{\theta}} \mathcal{J}_{\mathrm{RL}}(\boldsymbol{\theta})
        &=\mathbb{E}_{\boldsymbol{x} \sim D, \boldsymbol{y} \sim \pi_{\boldsymbol{\theta}}(\cdot \mid \boldsymbol{x})} \\
        &\quad\left[\sum_{t=1}^{T} A_{t} \nabla_{\boldsymbol{\theta}} \log \pi_{\boldsymbol{\theta}}\left(y_{t} \mid \boldsymbol{x}, \boldsymbol{y}_{<t}\right)\right],
    \end{aligned}
\end{equation}
where $A_{t}$ is the relative advantage of token $y_{t}$ over a baseline value.
Traditional methods such as PPO estimate $A_t$ via a learned critic, introducing substantial overhead. 
Group Relative Policy Optimization (GRPO) obviates the need for an additional value function approximator as required in PPO. Instead, it uses the average reward of multiple sampled outputs generated in response to the same question as the baseline. 
Formally, let $\pi_{\boldsymbol{{\theta}}}$ sample G responses $\{y_1, y_2, \cdots, y_G\}$ for
each prompt $x$.
The optimization objective of GRPO with token-
level loss and without the KL penalty term is:
\begin{equation}
\begin{aligned}
\label{GRPO}
\mathcal{J}_\text{GRPO}(\boldsymbol{\theta}) = & \mathbb{E}_{\boldsymbol{x} \sim D, \{\boldsymbol{y}^{(i)}\}_{i=1}^{G} \sim \pi_{\boldsymbol{\theta}_{\mathrm{old}}}(\cdot \mid \boldsymbol{x})} \\
& \left[\frac{1}{G}\sum_{i=1}^{G}\frac{1}{|y_{i}|}\sum_{t=1}^{|y_{i}|}\min \left(r_{i,t}(\boldsymbol{\theta}){A}_{i,t},\right.\right. \\
& \left.\left.\operatorname{clip}(r_{i,t}(\boldsymbol{\theta}),1-\varepsilon,1+\varepsilon){A}_{i,t}\right)\right],
\end{aligned}
\end{equation}
where $r_{i,t}(\boldsymbol{\theta})=\frac{\pi_{\boldsymbol{\theta}}(y_{i,t}|x,y_{i,<t})}{\pi_{\boldsymbol{\theta}_{\text{old}}}(y_{i,t}|x,y_{i,<t})}$, the advantage is given by ${A}_{i,t}=\frac{r(\boldsymbol{x}, \boldsymbol{y}^{(j)})-\mathrm{mean}\left(\left\{r(\boldsymbol{x}, \boldsymbol{y}^{(j)})\right\}_{j=1}^{G}\right)}{\mathrm{std}\left(\left\{r(\boldsymbol{x}, \boldsymbol{y}^{(j)})\right\}_{j=1}^{G}\right)}$, and $\varepsilon$ is a clipping hyperparameter.

\subsection{On-Policy Distillation}
The main idea of OPD is to let the student generate its own trajectories, and then minimize the reverse KL divergence between the student and the teacher on those student-generated trajectories:

\begin{equation}
  \begin{aligned}
    \mathcal{J}_{\mathrm{OPD}}(\boldsymbol{\theta})
    &= \min_{\boldsymbol{\theta}}\,
    \mathbb{E}_{\boldsymbol{x} \sim D,\, \boldsymbol{y} \sim \pi_{\boldsymbol{\theta}}(\cdot \mid \boldsymbol{x})} \\
    &\quad\left[
      \mathcal{D}_{\mathrm{KL}}\!\left(
        \pi_{\boldsymbol{\theta}}(\boldsymbol{y} \mid \boldsymbol{x})
        \,\|\,
        \pi^{*}(\boldsymbol{y} \mid \boldsymbol{x})
      \right)
    \right].
  \end{aligned}
\end{equation}

Then, we can get the gradient of OPD as:

\begin{equation}
\label{eq:opd_gradient}
  \begin{aligned}
        &\nabla_{\boldsymbol{\theta}} \mathcal{J}_{\mathrm{OPD}}(\boldsymbol{\theta})
        = \mathbb{E}_{\boldsymbol{x} \sim D,\, \boldsymbol{y} \sim \pi_{\boldsymbol{\theta}}(\cdot \mid \boldsymbol{x})} \\
        &\left[ \sum_{t=1}^{T} 
            \left( 
                \log \pi_{\boldsymbol{\theta}}\left(y_{t} \mid \boldsymbol{x}, \boldsymbol{y}_{<t}\right) 
                - \log \pi^{*}\left(y_{t} \mid \boldsymbol{x}, \boldsymbol{y}_{<t}\right) 
            \right) \right.\\
        &\quad \left. \cdot \nabla_{\boldsymbol{\theta}} \log \pi_{\boldsymbol{\theta}}\left(y_{t} \mid \boldsymbol{x}, \boldsymbol{y}_{<t}\right) 
        \right].
  \end{aligned}
\end{equation}

Eq.~\eqref{eq:opd_gradient} takes the same form as Eq.~\eqref{eq:grpo_gradient}, unifying OPD within the RL framework, where $A_{t}^\text{OPD}=-\left(\log \pi_{\boldsymbol{\theta}}\left(y_{t} \mid \boldsymbol{x}, \boldsymbol{y}_{<t}\right)-\log \pi^{*}\left(y_{t} \mid \boldsymbol{x}, \boldsymbol{y}_{<t}\right)\right)$ serves as the advantage in OPD, enabling token-level credit assignment. The detailed derivation is provided in Appendix~\ref{app:Preliminaries}.

\subsection{Supervised Fine-Tuning}
\label{sec:sft}
Supervised Fine-Tuning (SFT) trains the student policy $\pi_{\boldsymbol{\theta}}$ to mimic the teacher $\pi^{*}$ using demonstration data. From an information-theoretic perspective, SFT minimizes the forward KL divergence between the teacher and the student, which contrasts with the reverse KL divergence used in OPD:

\begin{equation}
  \begin{aligned}
    \mathcal{J}_{\mathrm{SFT}}(\boldsymbol{\theta})
    &= \min_{\boldsymbol{\theta}}\,
    \mathbb{E}_{\boldsymbol{x} \sim D} \\
    &\quad\left[
      \mathcal{D}_{\mathrm{KL}}\!\left(
        \pi^{*}(\boldsymbol{y} \mid \boldsymbol{x})
        \,\|\,
        \pi_{\boldsymbol{\theta}}(\boldsymbol{y} \mid \boldsymbol{x})
      \right)
    \right].
  \end{aligned}
\end{equation}

Ignoring the entropy of $\pi^{*}$ (independent of $\boldsymbol{\theta}$), this is equivalent to maximizing the expected log-likelihood of teacher trajectories:

\begin{equation}
    \max_{\boldsymbol{\theta}} \mathbb{E}_{\boldsymbol{x} \sim D, \boldsymbol{y} \sim \pi^{*}(\cdot \mid \boldsymbol{x})} \left[ \sum_{t=1}^{T} \log \pi_{\boldsymbol{\theta}}(y_t \mid \boldsymbol{x}, \boldsymbol{y}_{<t}) \right].
\end{equation}

Taking the gradient with respect to $\boldsymbol{\theta}$ yields:

\begin{equation}
\label{eq:sft_gradient}
  \begin{aligned}
    \nabla_{\boldsymbol{\theta}} \mathcal{J}_{\mathrm{SFT}}(\boldsymbol{\theta})
    &= \mathbb{E}_{\boldsymbol{x} \sim D,\, \boldsymbol{y} \sim \pi^{*}(\cdot \mid \boldsymbol{x})} \\
    &\quad\left[ \sum_{t=1}^{T} 1 \cdot \nabla_{\boldsymbol{\theta}} \log \pi_{\boldsymbol{\theta}}\left(y_{t} \mid \boldsymbol{x}, \boldsymbol{y}_{<t}\right) \right].
  \end{aligned}
\end{equation}

Comparing Eq.~\eqref{eq:sft_gradient} with the general policy gradient formulation in Eq.~\eqref{eq:grpo_gradient}, SFT can be elegantly unified within the RL framework as an \textit{off-policy} algorithm~\citep{DBLP:journals/corr/abs-2604-20244,DBLP:journals/corr/abs-2508-05629}. Specifically, the trajectories $\boldsymbol{y}$ are sampled from the teacher policy $\pi^{*}$ rather than the active student policy $\pi_{\boldsymbol{\theta}}$, and the advantage function is implicitly set to a constant $A_{t}^{\mathrm{SFT}} = 1$. This implies that every token generated by the teacher is treated as a gold standard, receiving a uniform, positive credit.

\begin{figure*}[t]
  \centering
  \includegraphics[width=\linewidth]{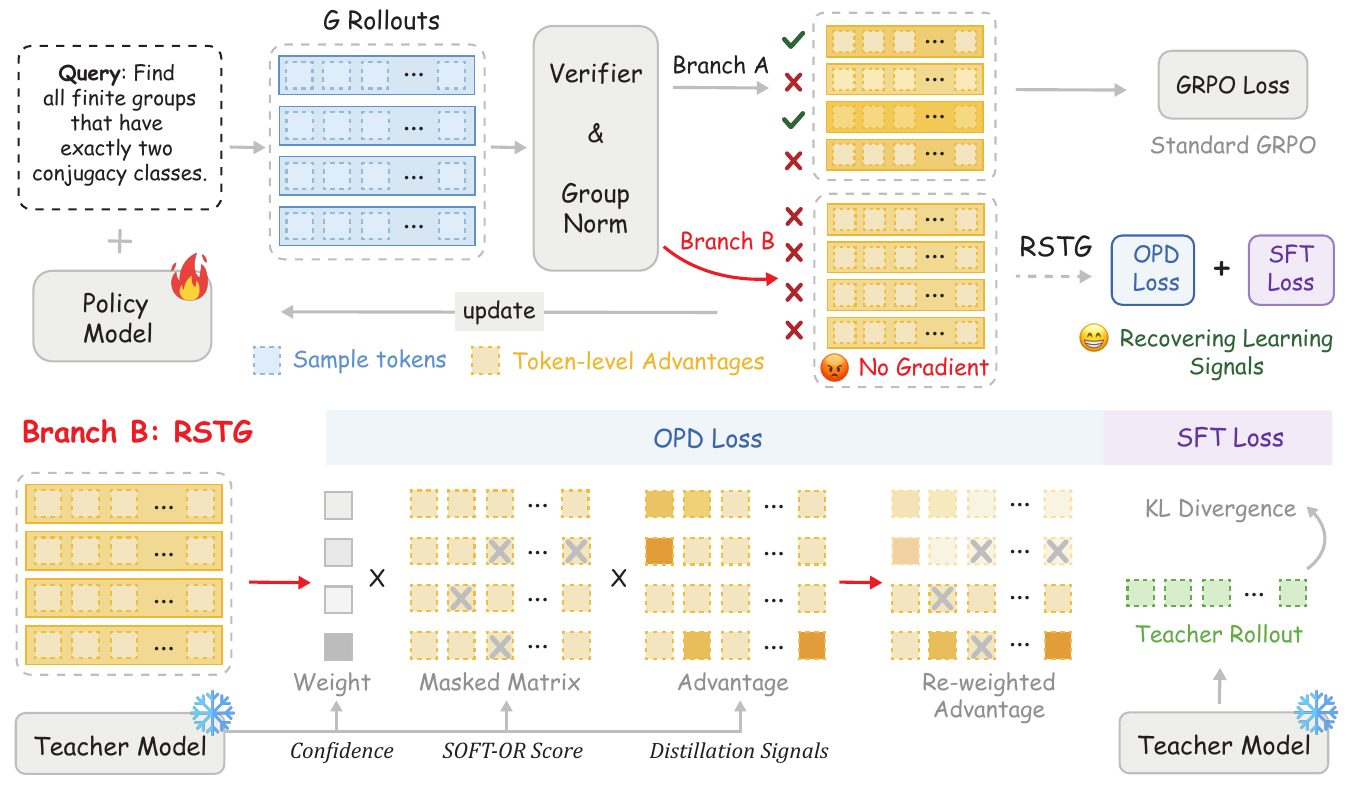}
  \caption{\textbf{Illustration of \methodname{} framework.} For negative zero-variance prompts, \methodname{} is applied; otherwise, standard GRPO is used. \methodname{} weights OPD advantages by the teacher's confidence, restricts gradient updates to high-quality tokens, and performs SFT on teacher-generated reference answers.}
  \label{fig:rstg_method}
\end{figure*}

\section{Method}
The framework of \methodname{} is illustrated in Figure~\ref{fig:rstg_method}. When a negative zero-variance prompt is encountered, the \methodname{} branch is activated; otherwise, the standard GRPO branch is followed. Within \methodname{}, we weight the OPD advantages according to the teacher's confidence on the given prompt (\S\ref{sec:method1}), restrict OPD gradient updates to high-quality tokens (\S\ref{sec:method2}), and perform SFT using reference answers generated by the teacher model (\S\ref{sec:method3}).

\subsection{Data Selection and Teacher-Guided Advantage Weighting}
\label{sec:method1}
\begin{figure}[t]
  \centering
  \includegraphics[width=\columnwidth]{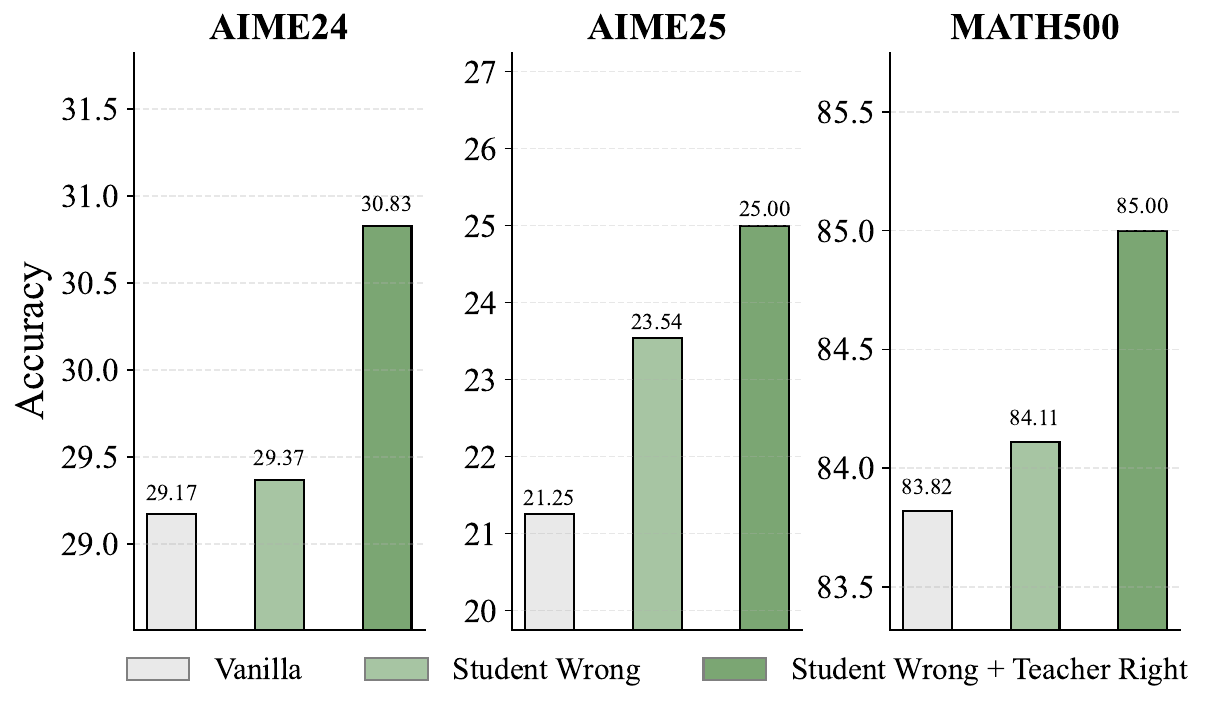}
  \caption{Performance on the mathmatical benchmark for Qwen3-1.7B-Instruct trained via OPD on different data partitions: Vanilla $\mathcal{D}$, Student Wrong $\mathcal{D}_{\text{sw}}$, Student Wrong and Teacher Right $\mathcal{D}_{\text{swtr}}$.}
  \label{fig:method1}
\end{figure}

To study the effects of data difficulty and teacher proficiency on OPD, we partition the full 57k-sample dataset $\mathcal{D}$ into two nested subsets based on model performance: (1) $\mathcal{D}_{\text{sw}}$ (9k samples), where the student fails all 8 rollouts ($\text{mean@8} = 0$), and (2) $\mathcal{D}_{\text{swtr}}$ (2k samples), a subset of $\mathcal{D}_{\text{sw}}$ where the teacher achieves a perfect success rate ($\text{mean@8} = 1$). We evaluate OPD across these three data scales on mathematical benchmarks (Figure~\ref{fig:method1}). The results yield two key insights:

\begin{tcolorbox}[colback=green!5, colframe=green!40, boxrule=0.4pt, arc=1.5pt, left=4pt, right=4pt, top=2pt, bottom=2pt]
\textbf{Finding 1.} Training on tasks challenging for the student ($\mathcal{D}_{\text{sw}}$) outperforms training on the full dataset ($\mathcal{D}$).
\end{tcolorbox}

\begin{tcolorbox}[colback=green!5, colframe=green!40, boxrule=0.4pt, arc=1.5pt, left=4pt, right=4pt, top=2pt, bottom=2pt]
\textbf{Finding 2.} Restricting training to samples where the teacher excels ($\mathcal{D}_{\text{swtr}}$) yields the best performance using only 3.63\% of $\mathcal{D}$.
\end{tcolorbox}

This indicates that the complementary effect of OPD on RL should be channeled specifically into this category of samples, which naturally correspond to \textit{negative zero-variance prompts} in GRPO, where all rollouts fail and the advantage collapses to zero, leaving no training signal. OPD effectively recovers these lost gradient signals by providing token-level advantages. Furthermore, we incorporate teacher confidence as a fine-grained weighting mechanism into this hybrid objective.

Formally, for sample $i$, let $\omega_i \in [0, 1]$ denote the teacher's $\text{mean@8}$ score, serving as a proxy for teacher confidence to dynamically scale the token-level OPD advantage. The hybrid advantage function is defined as:

\begin{equation}
    {A}_{i, t}^{\text{Hybrid}} = 
    \begin{cases} 
        \beta \cdot \omega_i \cdot {A}_{i, t}^{\text{OPD}}, & \text{if } r_j = 0, \ \forall j \in G \\ 
        {{A}}_{i, t}^{\text{GRPO}}, & \text{otherwise}
    \end{cases}
    \label{eq:hybrid_advantage}
\end{equation}

where ${A}_{i, t}^{\text{GRPO}}$ denotes the standard GRPO advantage, ${A}_{i, t}^{\text{OPD}}$ the OPD advantage, and $\beta$ a scaling hyperparameter controlling the magnitude of the OPD signal. The weighting coefficient $\omega_i$ assigns greater optimization weight to tokens where the teacher is more confident, leading to more reliable policy optimization.

\subsection{Mitigating Premature Convergence via Token Selection}
\label{sec:method2}

We select high-quality tokens to slow the student's convergence toward the teacher and reduce gradient noise. Inspired by TIP~\citep{xu2026tip}, we identify two categories of valuable tokens: (1) \textbf{high-entropy tokens}, where the student is uncertain, typically representing critical reasoning junctures; and (2) \textbf{tokens with large teacher-student discrepancy}, which tend to carry richer information.

We define the student's entropy at position $t$ as:
\begin{equation}
    h_{t}=-\sum_v p_{t,v} \log p_{t,v},
    \label{eq:ht}
\end{equation}
where $p_{t,v}$ is the student's predicted probability for vocabulary token $v$ at position $t$. The teacher-student discrepancy is defined as:
\begin{equation}
    d_{t}=|{A}_{i, t}^{\text{OPD}}|.
    \label{eq:dt}
\end{equation}

Tokens with large $h_{t}$ or $d_{t}$ are considered high-value. We combine these two criteria via the Soft-OR formulation (see Appendix~\ref{app:Method2} for details). Given min-max normalized $\hat{h}_{t}, \hat{d}_{t} \in [0, 1]$, the selection score is:
\begin{equation}
    s_{t}= \hat{h}_{t} + \hat{d}_{t} - \hat{h}_{t} \cdot \hat{d}_{t}.
    \label{eq:st}
\end{equation}

The token-masked OPD advantage is then defined as:
\begin{equation}
    \hat{A}_{i,t}^{\text{OPD}} = 
    \begin{cases} 
        {A}_{i,t}^{\text{OPD}}, & \text{if } t \in \mathcal{S}_k \\[4pt]
        0, & \text{otherwise}
    \end{cases}
    \label{eq:masked_opd}
\end{equation}
where $\mathcal{S}_k$ denotes the top-$k\%$ tokens ranked by $s_t$, enabling gradual distillation of the teacher's capabilities while reducing gradient noise.

\subsection{Complementary Supervision on Negative Zero-Variance Prompts}
\label{sec:method3}
\begin{figure}[t]
  \centering
  \includegraphics[width=\columnwidth]{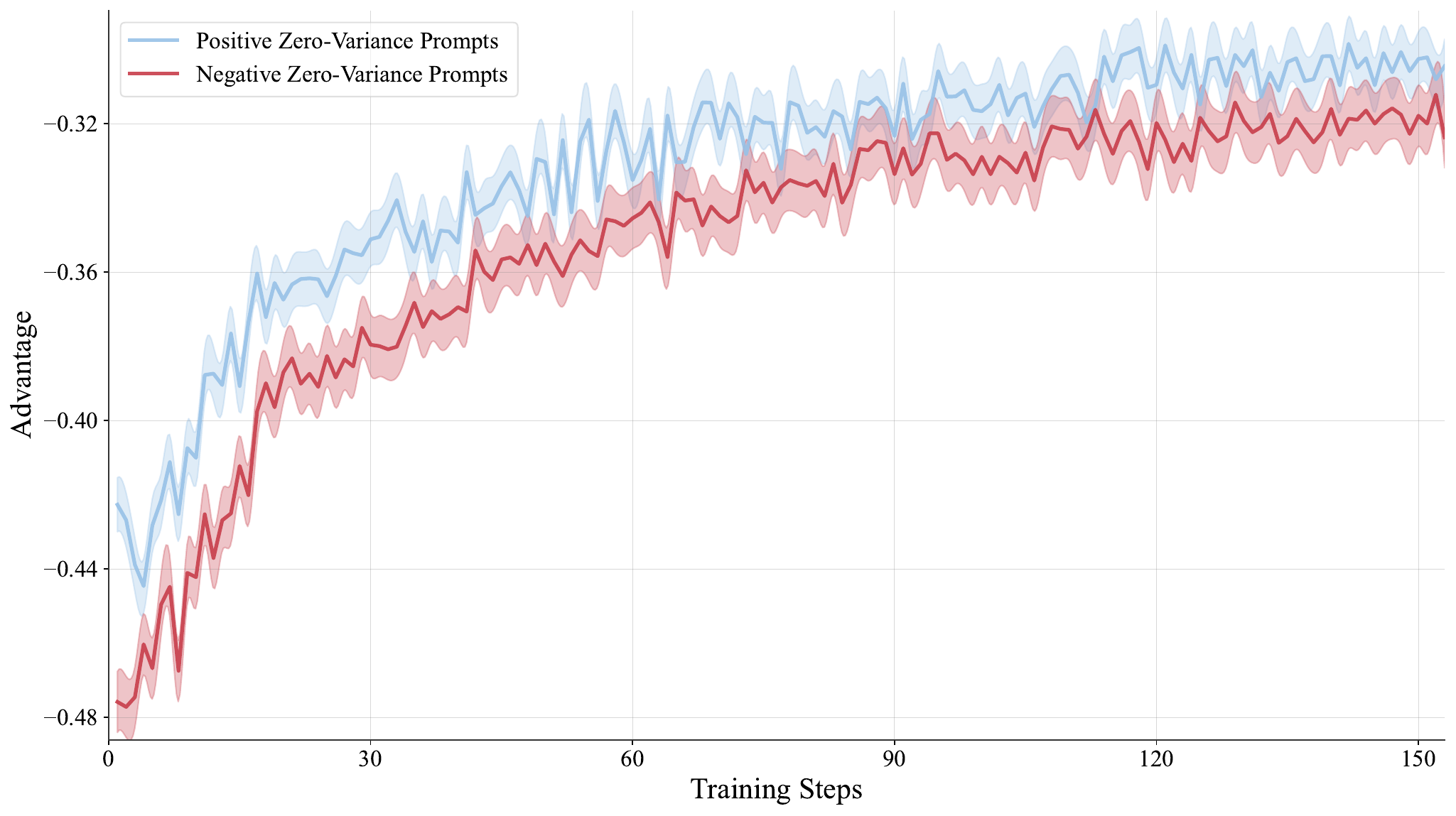}
  \caption{OPD advantage curves during training on positive and negative zero-variance prompts.}
  \label{fig:advantage}
\end{figure}

This advantage asymmetry is further exacerbated on negative zero-variance prompts. As illustrated in Figure~\ref{fig:advantage}, the advantage values on negative zero-variance prompts are consistently more negative than those on positive zero-variance prompts throughout training, reflecting a severely suppressive optimization signal that 
penalizes the student without providing any constructive guidance toward correct solutions. 


To mitigate this advantage asymmetry of OPD, we augment the training objective for negative zero-variance prompts 
with an auxiliary SFT loss computed on correct trajectories pre-generated out by the teacher $\pi^{*}$. 
Formally, the final training objective is defined as:

\begin{equation}
\scalebox{0.9}{$
    \mathcal{J}_{\text{Final}}(\boldsymbol{\theta}) = 
    \begin{cases} 
        \mathcal{J}_{\text{GRPO}}\left(\boldsymbol{\theta}; \beta \cdot \omega_i \cdot \hat{A}_{i,t}^{\text{OPD}}\right) \\
        + \beta \cdot \mathcal{J}_{\text{SFT}}(\boldsymbol{\theta}), \quad \text{if } r_j = 0, \ \forall j \in G \\[6pt]
        \mathcal{J}_{\text{GRPO}}\left(\boldsymbol{\theta}; {{A}}_{i,t}^{\text{GRPO}}\right), \quad\text{otherwise}
    \end{cases}
$}
    \label{eq:final_objective}
\end{equation}

where $\mathcal{J}_{\text{SFT}}$ is computed exclusively on negative zero-variance prompts, and $\beta$ is shared with the OPD term in Eq.~\eqref{eq:hybrid_advantage}, ensuring consistent regularization across both components.

As established in Section~\ref{sec:sft}, this objective is equivalent to assigning a uniform positive advantage $A_t^{\text{SFT}} = 1$ to every token in the teacher trajectory, which injects purely positive gradient signals into the optimization on prompts where RL yields no learning signal, thereby shifting the OPD advantage space toward the positive direction, alleviating the advantage asymmetry issue, while simultaneously introducing a global perspective signal and guiding the student toward correct solution trajectories demonstrated by the teacher.

\section{Experiments}

\definecolor{cvprblue}{rgb}{0.21,0.49,0.74}

\begin{table*}[t]
    \centering
    \footnotesize
    \caption{Main results on MATH and CODE benchmarks across three Student-Teacher model pairs. \textbf{Best} and \underline{second-best} are highlighted.}
    \renewcommand{\arraystretch}{1.1}
    \setlength{\tabcolsep}{6pt}
    \begin{tabular}{l ccccc ccc}
        \toprule
        \multirow{2}{*}[-4pt]{Method}
        & \multicolumn{5}{c}{\textbf{MATH}}
        & \multicolumn{3}{c}{\textbf{CODE}} \\
        \cmidrule(lr){2-6} \cmidrule(l){7-9}
        & AIME24 & AIME25 & MATH500 & OLMPIAD & AVG
        & APPS & MBPP+ & AVG \\
        \midrule
        \rowcolor{gray!10} \multicolumn{9}{l}{\textit{Student: Qwen3-1.7B-Instruct \quad Teacher: Qwen3-4B-Instruct-2507}} \\
        Vanilla      & 12.08 & 10.83 & 72.66 & 40.26 & 39.96 & 30.80 & 49.22 & 40.01 \\
        OPD          & 32.29 & 20.62 & 84.32 & 53.46 & 47.67 & 32.41 & \textbf{75.10} & 53.76 \\
        GRPO         & 34.79 & 27.71 & 88.64 & 55.15 & 51.57 & 50.39 & 69.66 & 60.03 \\
        GRPO+OPD    & \underline{35.21} & 26.88 & 88.27 & 55.13 & 51.37 & 57.03 & \underline{72.06} & \underline{64.55} \\
        ReLIFT     & \underline{35.21} & \underline{30.42} & \underline{88.65} & \underline{55.79} & \underline{52.52} & \underline{60.17} & 60.98 & 60.58 \\
        RL-ZVP       & 31.87 & 25.62 & 86.99 & 53.46 & 49.49 & 49.73 & 70.12 & 59.93 \\
        \textbf{\methodname{}} & \textbf{42.98} & \textbf{31.87} & \textbf{89.36} & \textbf{57.36} & \textbf{55.39} & \textbf{61.81} & \underline{72.41} & \textbf{67.11} \\
        \midrule
        \rowcolor{gray!10} \multicolumn{9}{l}{\textit{Student: Qwen3-4B-Instruct \quad Teacher: Qwen3-4B-Instruct-2507}} \\
        Vanilla      & 24.38 & 18.33 & 83.84 & 52.20 & 44.69 & 43.93 & 74.22 & 59.08 \\
        OPD          & 50.83 & 40.63 & 91.93 & 63.71 & 61.78 & 53.10 & \underline{86.19} & 69.65 \\
        GRPO         & 54.37 & \underline{45.00} & 94.29 & 66.08 & 64.94 & 65.29 & 77.82 & 71.56 \\
        GRPO+OPD    & 50.42 & 41.87 & 93.95 & \textbf{67.25} & 63.37 & 73.46 & 84.53 & 79.00 \\
        ReLIFT     & \textbf{58.13} & 42.71 & \underline{94.74} & 64.61 & \underline{65.05} & \textbf{76.12} & 85.41 & \underline{80.77} \\
        RL-ZVP       & 50.42 & 42.71 & 93.54 & 63.41 & 62.52 & 68.16 & 82.98 & 75.57 \\
        \textbf{\methodname{}} & \underline{57.08} & \textbf{47.92} & \textbf{95.30} & \underline{67.24} & \textbf{66.89} & \underline{75.68} & \textbf{88.42} & \textbf{82.05} \\
        \midrule
        \rowcolor{gray!10} \multicolumn{9}{l}{\textit{Student: Qwen2.5-3B-Instruct \quad Teacher: Qwen2.5-14B-Instruct}} \\
        Vanilla      & 4.58 & 1.25 & 62.14 & 27.04 & 23.75 & 16.60 & 62.84 & 39.72 \\
        OPD          & 6.67 & 1.46 & 63.56 & 28.98 & 25.17 & 23.14 & 66.63 & 44.89 \\
        GRPO         & 6.04 & 3.33 & \textbf{68.16} & \underline{32.09} & 27.41 & 37.18 & 66.63 & 51.91 \\
        GRPO+OPD    & \underline{6.88} & \underline{4.58} & 67.68 & 31.95 & \underline{27.77} & 49.72 & \underline{73.37} & 61.55 \\
        ReLIFT     & 6.46 & 2.71 & 67.44 & 32.05 & 27.17 & \underline{52.60} & 71.50 & \underline{62.05} \\
        RL-ZVP       & 6.46 & 3.33 & 67.24 & 31.72 & 27.19 & 42.03 & 71.01 & 57.52 \\
        \textbf{\methodname{}} & \textbf{8.33} & \textbf{5.42} & \underline{68.14} & \textbf{32.74} & \textbf{28.66} & \textbf{53.01} & \textbf{74.03} & \textbf{63.52} \\
        \bottomrule
    \end{tabular}
    \label{tab:main_result}
\end{table*}
\definecolor{tealcolor}{RGB}{89, 167, 168}
\definecolor{lightcyan}{RGB}{224, 242, 241}

\begin{figure*}[t]

  \includegraphics[width=0.32\linewidth]{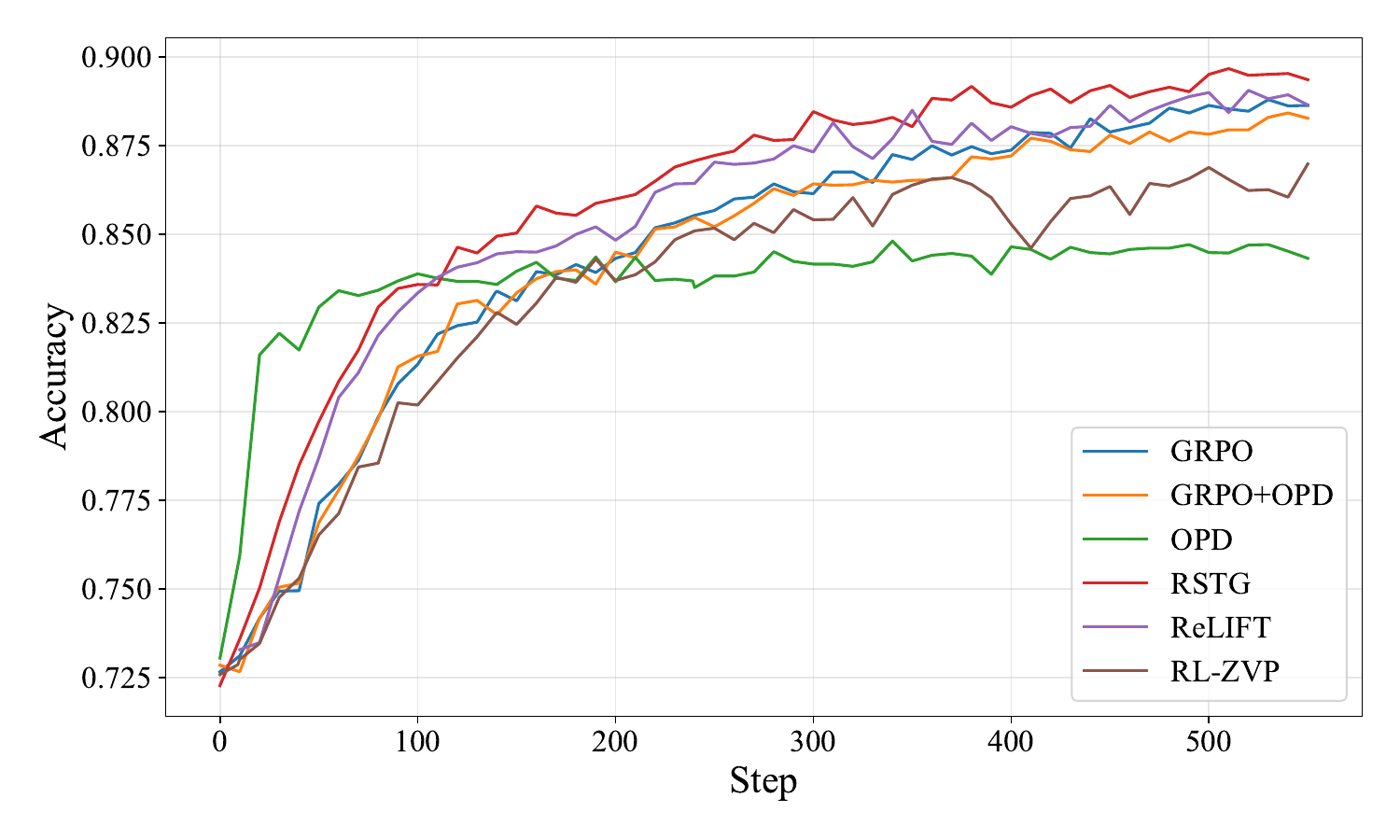} \hfill
 \includegraphics[width=0.32\linewidth]{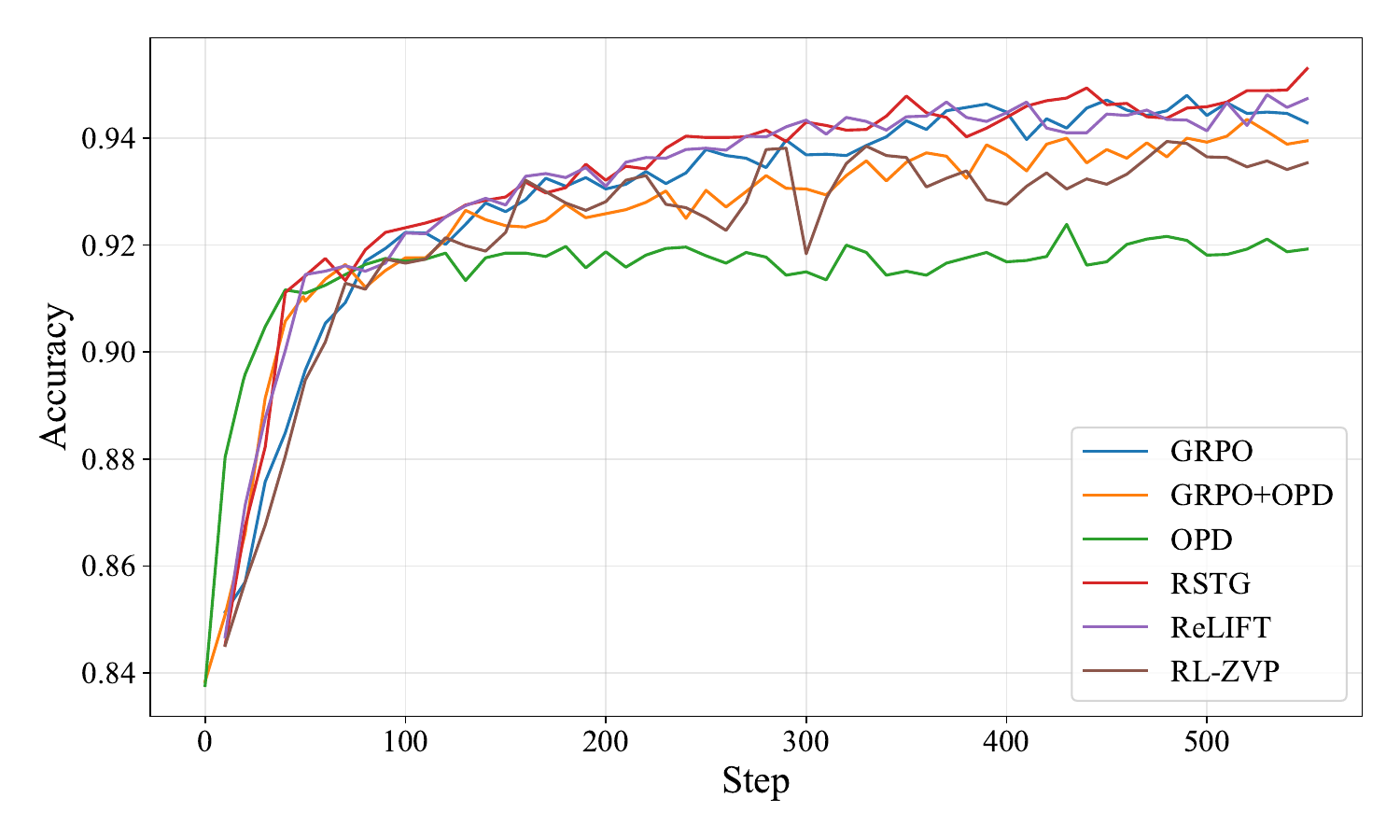} \hfill
 \includegraphics[width=0.32\linewidth]{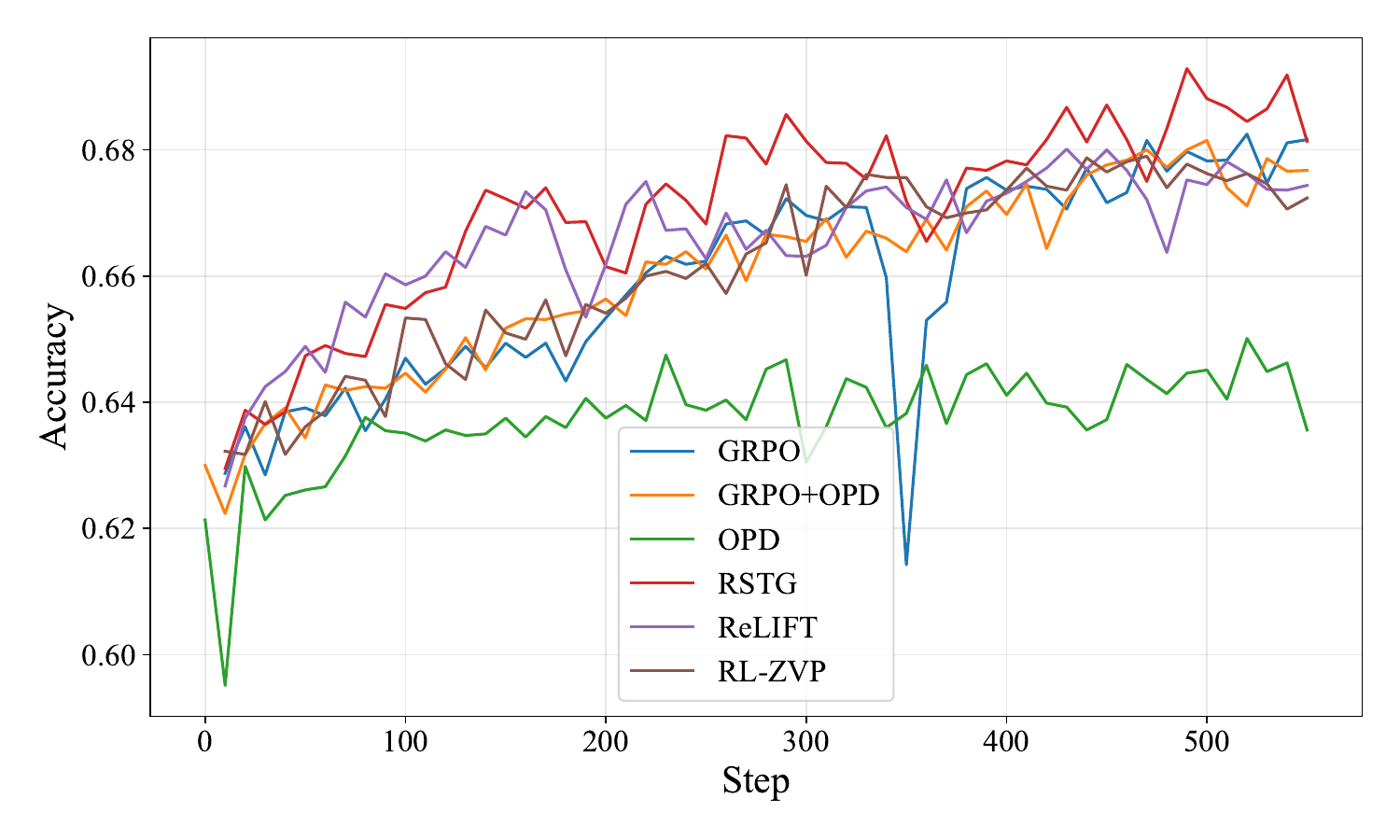} 
  \caption {Performance on MATH500 for Qwen3-1.7B-Instruct $\rightarrow$ Qwen3-4B-Instruct-2507, Qwen3-4B-Instruct $\rightarrow$ Qwen3-4B-Instruct-2507, and Qwen2.5-3B-Instruct $\rightarrow$ Qwen2.5-14B-Instruct from left to right, respectively.}
  \label{fig:main_results}
\end{figure*}

\subsection{Setup}

\paragraph{Models}
We employ three teacher-student model pairs across two model families:
\begin{itemize}[itemsep=0pt, leftmargin=*]
    \item \textbf{Qwen3}~\citep{DBLP:journals/corr/abs-2505-09388}: Qwen3-1.7B-Instruct $\rightarrow$ Qwen3-4B-Instruct-2507 (Pair \ding{182}); Qwen3-4B-Instruct $\rightarrow$ Qwen3-4B-Instruct-2507 (Pair \ding{183}).
    \item \textbf{Qwen2.5}~\citep{DBLP:journals/corr/abs-2412-15115}: Qwen2.5-3B-Instruct $\rightarrow$ Qwen2.5-14B-Instruct (Pair \ding{184}).
\end{itemize}

\paragraph{Dataset and Metrics.}
For training, we filter the DeepMath~\citep{DBLP:journals/corr/abs-2504-11456} dataset to select 57K samples with difficulty $\geq 6$ as math RL data, and use Eurus-RL-Code~\citep{DBLP:journals/corr/abs-2502-01456} (25K samples) as code RL data. For evaluation, we use AIME 2024~\citep{aime24}, AIME 2025~\citep{aime25}, MATH-500~\citep{DBLP:conf/iclr/Gao0YCMDLMCXTWZ25}, and OLMPIAD Bench~\citep{DBLP:conf/acl/HeLBHTSHHHZLQL024} for mathematical reasoning, and APPS~\citep{DBLP:conf/nips/HendrycksBKMAGB21} and MBPP+~\citep{DBLP:conf/nips/LiuXW023} for code generation. We set temperature to 1.0, top-p to 1.0, and maximum generation length to 8,192, sampling 16 solutions per math problem and 4 per code problem. Further details are in Appendix~\ref{app:Experiment Details}.

\paragraph{Implementation Details.}
We set the batch size to 256, maximum response length to 8192, number of rollouts to 8, and learning rate to $1\times10^{-6}$, with thinking mode disabled for all models. For mathematics, training runs for 550 steps (222 steps/epoch); for code, 400 steps (98 steps/epoch), both reaching convergence. The coefficient $\beta$ in Eq.~\ref{eq:final_objective} is linearly annealed with $\beta_{\text{init}} = 5 \times 10^{-3}$, $\delta = 5 \times 10^{-5}$, and $\beta_{\text{min}} = 1 \times 10^{-3}$. To construct the SFT dataset, we pre-sample $n=8$ responses per prompt from the teacher, retaining the shortest correct response as reference. More details and cost analysis are provided in Appendix~\ref{app:Implementation Details.}.

\paragraph{Baselines}
We compare our method against five baselines: GRPO, OPD, 
\textbf{GRPO+OPD}, which directly combines the GRPO and OPD losses following the same configuration as KDRL~\citep{DBLP:journals/corr/abs-2506-02208}. See Appendix~\ref{app:Baselines} for more details. Since our method specifically targets negative zero-variance prompts, we also include two representative baselines from the line of work on learning from negatives: \textbf{ReLIFT}~\citep{DBLP:journals/corr/abs-2506-07527}, which applies SFT using ground-truth answers on negative zero-variance prompts, and \textbf{RL-ZVP}~\citep{DBLP:journals/corr/abs-2509-21880}, which designs an asymmetric advantage formulation based on token-level entropy for both positive and negative zero-variance prompts to extract effective learning signals. As our primary objective is to push the upper bound of reinforcement learning, all baselines above are built upon GRPO, with the exception of OPD.

\subsection{Main Results}

Table~\ref{tab:main_result} presents the performance of all methods across four mathematical reasoning benchmarks and two code generation benchmarks. 

For mathematics, naive GRPO+OPD tends to slightly underperform standard GRPO, with changes of $-0.2\%$, $-1.57\%$, and $+0.36\%$ across the three model pairs. In contrast, \methodname{} achieves substantial improvements over naive GRPO+OPD, with gains of $+4.02\%$, $+3.52\%$, and $+0.89\%$ respectively, surpassing nearly all baselines.

For code generation, naive GRPO+OPD yields notable improvements over standard GRPO, with gains of $+4.52\%$, $+7.44\%$, and $+9.64\%$ across the three model pairs. Nevertheless, \methodname{} achieves consistent further improvements over naive GRPO+OPD, with gains of $+2.56\%$, $+3.05\%$, and $+1.97\%$ across the three model pairs, surpassing almost all baselines.

Additionally, we observe that ReLIFT achieves relatively strong performance among the baselines, as directly supervising the model with ground-truth answers is effective for inherently challenging negative zero-variance prompts. However, it is confined to offline imitation learning. OPD complements this by providing online corrective signals at every token generation step, and with our carefully designed components (\S\ref{para: opd design}), the two objectives are organically integrated to reinforce each other, leading to more effective learning.

Figure~\ref{fig:main_results} illustrates the training dynamics of all methods on MATH500 as a representative benchmark, where \methodname{} outperforms nearly all baselines at every training step. Notably, for Model Pair 2, where both the teacher and student are 4B models with a relatively small capability gap, the performance gains are less pronounced compared to the other two model pairs. More complete training curves are provided in the Appendix~\ref{app:Training Dynamics}.

\subsection{The Asymmetry of Advantage}

\label{sec:The Symmetry of Advantage}
\begin{figure}[t]
  \includegraphics[width=\columnwidth]{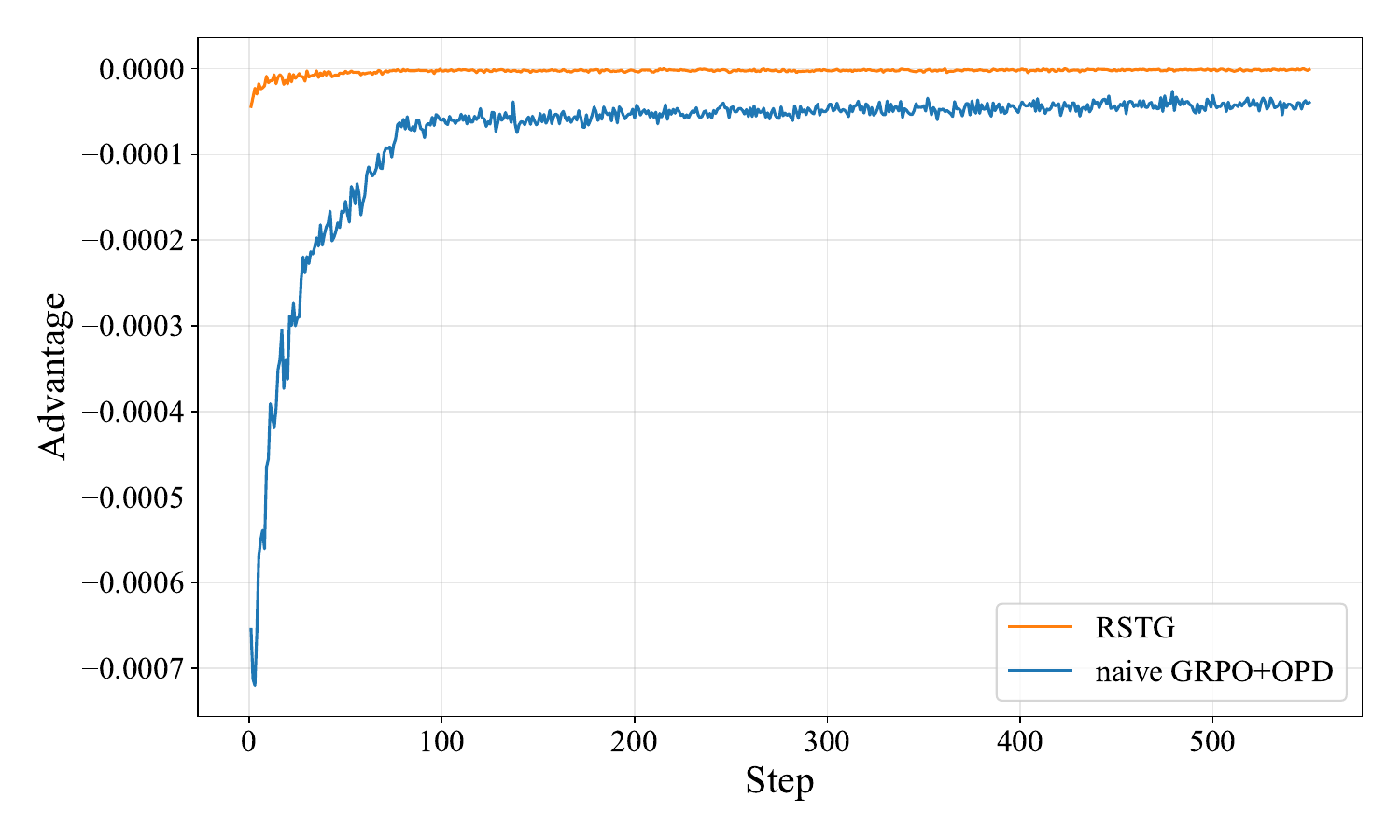}
  \caption{Comparison of advantage values during training between \methodname{} and naive GRPO+OPD, using model pair 2 as a representative example.}
  \label{fig:adv}
\end{figure}

We compared the evolution of advantages during training between \methodname{} and naive GRPO+OPD. As shown in Figure~\ref{fig:adv}, \methodname{} yields higher advantage values and alleviates the advantage asymmetry introduced by OPD, thereby allowing more tokens to receive positive learning signals.

\subsection{Convergence Speed and Response Length}

\paragraph{Convergence Speed.} We use the overlap $M_{\text{overlap}}$ between the top-$k$ tokens ($k=16$) of the teacher and the student~\citep{DBLP:journals/corr/abs-2604-13016} to quantify the alignment between their candidate spaces, where a higher value indicates that the student more closely fits the teacher. We monitor this metric to track the rate at which the student converges to the teacher. At step 200, standard OPD reaches $69.7\%$, naive GRPO+OPD reaches $67.6\%$, while \methodname{} achieves only $65.81\%$, demonstrating that \methodname{} slows convergence toward the teacher and preserves a larger exploration space for RL.

\begin{figure}[t]
  \includegraphics[width=\columnwidth]{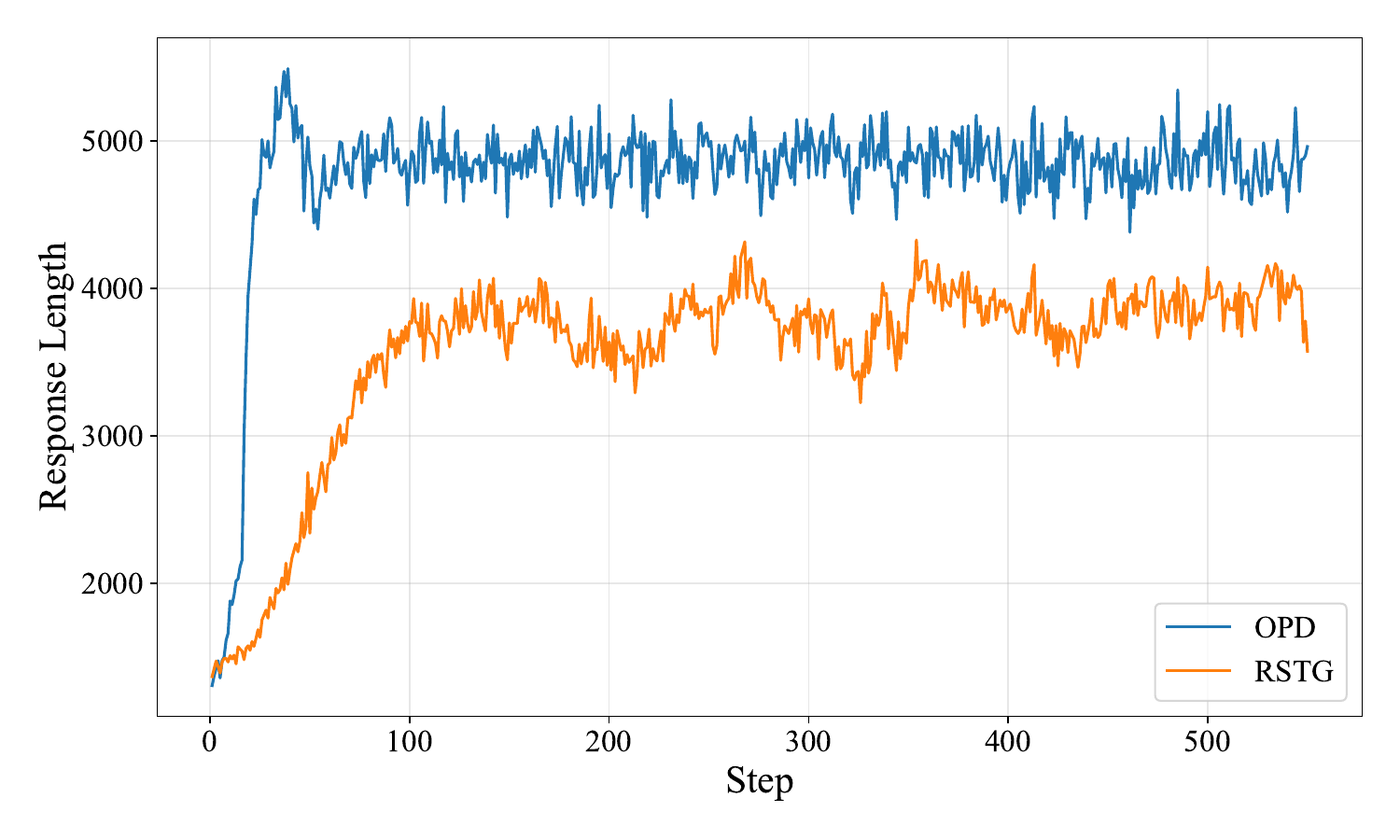}
  \caption{Comparison of response length during training between \methodname{} and OPD.}
  \label{fig:len}
\end{figure}

\paragraph{Response Length.} OPD suffers from abrupt length inflation~\citep{DBLP:journals/corr/abs-2604-08527,fu2026revisiting}, as illustrated in Figure~\ref{fig:len}. Our \methodname{} effectively mitigates this issue, maintaining response length comparable to that of GRPO.

\begin{figure}[t]
  \includegraphics[width=\columnwidth]{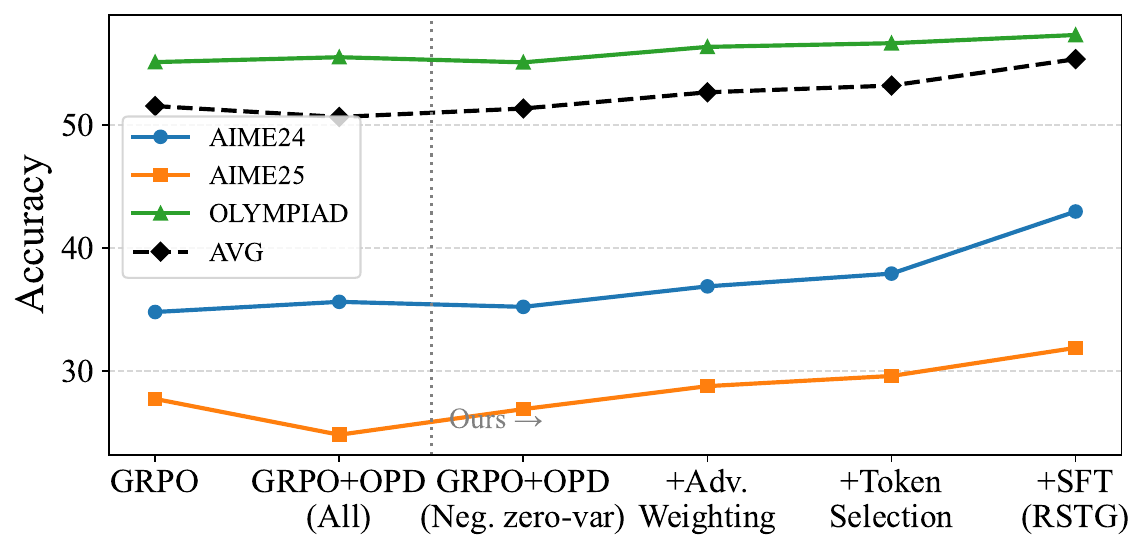}
  \caption{Performance on mathematical benchmarks as components of \methodname{} are progressively added. MATH500 is excluded from the figure due to its larger score range, but is included in the AVG computation.}
  \label{fig:ablation}
\end{figure}

\begin{table}[h]
\centering
\caption{Average accuracy on mathematical benchmarks under different $\beta$ strategies. Best values are \textbf{bolded}.}
\label{tab:beta}
\resizebox{\columnwidth}{!}{%
\begin{tabular}{lccccc}
\toprule
{\small \textbf{Method}} & {\small \textbf{AIME24}} & {\small \textbf{AIME25}} & {\footnotesize \textbf{MATH500}} & {\footnotesize \textbf{OLMPIAD}} & {\textbf{AVG}} \\
\midrule
\rowcolor{gray!10} $\beta_{\text{init}}=5\times10^{-3}$ & 35.21 & \textbf{26.88} & \textbf{88.27} & \textbf{55.13} & \textbf{51.37} \\
$\beta_{\text{init}}=1\times10^{-2}$ & \textbf{36.04} & 26.46 & 87.35 & 54.01 & 50.97 \\
$\beta=5\times10^{-3}$                      & 35.00 & 25.21 & 87.40 & 53.32 & 50.23 \\
$\beta=5\times10^{-2}$                   & 34.79 & 24.58 & 86.55 & 52.18 & 49.53 \\
$\beta=1\times10^{-1}$                       & 32.08 & 24.38 & 85.81 & 51.51 & 48.45 \\
\bottomrule
\end{tabular}%
}
\end{table}
\subsection{Ablation Studies}
\paragraph{Each component plays a crucial role.}
Taking model pair1 as an example, we start from a baseline that naively combines the GRPO and OPD losses over all samples, and progressively incorporate each component of our \methodname{}. As shown in Figure~\ref{fig:ablation}, each component contributes meaningfully to the overall performance. Detailed results are provided in the Appendix~\ref{app:Ablation Studies}.

\paragraph{The careful design of OPD is indispensable.}
\label{para: opd design}
When the carefully designed OPD components are removed from \methodname{}, specifically teacher-guided advantage weighting and token selection, the average accuracy on the math benchmarks for Pair 1 drops to 51.74\%, which is even lower than applying SFT on top of GRPO alone (i.e., ReLIFT in Table~\ref{tab:main_result}, 52.52\%). This outcome is expected: without these carefully designed components, OPD fails to operate on the appropriate samples, and the absence of token selection causes the model to converge prematurely toward the teacher distribution, thereby undermining its exploratory capacity during RL training. This demonstrates that our OPD design choices are indispensable. Only with these designs in place can OPD and SFT be effectively integrated, enabling the model to simultaneously imitate the teacher's correct responses via SFT and receive fine-grained corrective signals at every token generation step via OPD, thus achieving more effective and efficient training. Without such elaborate design, OPD may even exert a detrimental effect on overall performance.

\paragraph{The coefficient $\beta$ of OPD and SFT.}

We investigate two strategies for $\beta$: constant and linear annealing. For the constant strategy, we evaluate several fixed values of $\beta$; for the linear annealing strategy, we evaluate several different initial values $\beta_{\text{init}}$. In both cases, OPD is applied solely on negative zero-variance prompts without any additional components. Results are reported in Table~\ref{tab:beta}.

A constant $\beta$ consistently degrades performance, as the performance upper bound of OPD is lower than that of GRPO. A large $\beta$ causes the model to overfit to the teacher and converge prematurely to this lower bound. The linear annealing strategy suffers from the same issue when the initial value is too large, while too small an initial value results in a distillation signal too weak to be effective. We therefore adopted $\beta_{\text{init}} = 5 \times 10^{-3}$ as our final choice. Since the SFT intensity is aligned with that of OPD, both terms share the same coefficient $\beta$.

\section{Conclusion}
To address the performance degradation caused by the naive combination of GRPO and OPD, we propose \methodname{}, which selectively applies OPD on difficult samples that the student has not yet mastered, weighted by the teacher's degree of proficiency. Furthermore, \methodname{} employs fine-grained token selection to slow down convergence and reduce gradient noise, and incorporates an auxiliary SFT objective to inject positive gradient signals and provide a global perspective on the solution space. Experiments across benchmarks in two domains and three model pairs confirm consistent improvements over both standard GRPO and naive GRPO+OPD baselines.

\section*{Limitations}
We conduct experiments across two domains, math and code, which demonstrate the generalizability of our method. Nevertheless, we look forward to exploring its applicability in other settings, such as agentic tasks. Due to computational constraints and the strict requirement of OPD for a meaningful capability gap between the teacher and student, our experiments are conducted at the largest scale feasible within our resources. We expect \methodname{} to remain effective at larger model scales, and leave this for future work.

\bibliography{custom}
\clearpage
\appendix

\section{Appendix}
\label{sec:appendix}

\subsection{Preliminaries}
\label{app:Preliminaries}

The main idea of OPD is to let the student generate its own trajectories, and then minimize the reverse KL divergence between the student and the teacher on those student-generated trajectories:

\begin{equation}
  \begin{aligned}
    \mathcal{J}_{\mathrm{OPD}}(\boldsymbol{\theta})
    &= \min_{\boldsymbol{\theta}}\,
    \mathbb{E}_{\boldsymbol{x} \sim D,\, \boldsymbol{y} \sim \pi_{\boldsymbol{\theta}}(\cdot \mid \boldsymbol{x})} \\
    &\quad\left[
      \mathcal{D}_{\mathrm{KL}}\!\left(
        \pi_{\boldsymbol{\theta}}(\boldsymbol{y} \mid \boldsymbol{x})
        \,\|\,
        \pi^{*}(\boldsymbol{y} \mid \boldsymbol{x})
      \right)
    \right].
  \end{aligned}
\end{equation}

Expanding the KL divergence and applying the chain rule of probability, we have:

\begin{equation}
  \begin{aligned}
    \mathcal{J}_{\mathrm{OPD}}(\boldsymbol{\theta})
    &= \mathbb{E}_{\boldsymbol{x},\, \boldsymbol{y} \sim \pi_{\boldsymbol{\theta}}}
    \left[
      \log \frac{\pi_{\boldsymbol{\theta}}(\boldsymbol{y} \mid \boldsymbol{x})}{\pi^{*}(\boldsymbol{y} \mid \boldsymbol{x})}
    \right] \\
    &= \mathbb{E}_{\boldsymbol{x},\, \boldsymbol{y} \sim \pi_{\boldsymbol{\theta}}}
    \left[
      \sum_{t=1}^{T}
      \log \frac{\pi_{\boldsymbol{\theta}}(y_t \mid \boldsymbol{x}, \boldsymbol{y}_{<t})}
               {\pi^{*}(y_t \mid \boldsymbol{x}, \boldsymbol{y}_{<t})}
    \right].
  \end{aligned}
\end{equation}

Taking the gradient with respect to $\boldsymbol{\theta}$ and applying the log-derivative trick $\nabla_{\boldsymbol{\theta}} \mathbb{E}_{\boldsymbol{y} \sim \pi_{\boldsymbol{\theta}}}[f(\boldsymbol{y})] = \mathbb{E}_{\boldsymbol{y} \sim \pi_{\boldsymbol{\theta}}}[f(\boldsymbol{y}) \nabla_{\boldsymbol{\theta}} \log \pi_{\boldsymbol{\theta}}(\boldsymbol{y})]$:

\begin{equation}
\label{eq:opd_gradient}
  \begin{aligned}
        &\nabla_{\boldsymbol{\theta}} \mathcal{J}_{\mathrm{OPD}}(\boldsymbol{\theta})
        = \mathbb{E}_{\boldsymbol{x} \sim D,\, \boldsymbol{y} \sim \pi_{\boldsymbol{\theta}}(\cdot \mid \boldsymbol{x})} \\
        &\left[ \sum_{t=1}^{T} 
            \left( 
                \log \pi_{\boldsymbol{\theta}}\left(y_{t} \mid \boldsymbol{x}, \boldsymbol{y}_{<t}\right) 
                - \log \pi^{*}\left(y_{t} \mid \boldsymbol{x}, \boldsymbol{y}_{<t}\right) 
            \right) \right.\\
        &\quad \left. \cdot \nabla_{\boldsymbol{\theta}} \log \pi_{\boldsymbol{\theta}}\left(y_{t} \mid \boldsymbol{x}, \boldsymbol{y}_{<t}\right) 
        \right].
  \end{aligned}
\end{equation}

Eq.~\eqref{eq:opd_gradient} takes the same form as Eq.~\eqref{eq:grpo_gradient}, unifying OPD within the RL framework, where $A_{t}^\text{OPD}=-\left(\log \pi_{\boldsymbol{\theta}}\left(y_{t} \mid \boldsymbol{x}, \boldsymbol{y}_{<t}\right)-\log \pi^{*}\left(y_{t} \mid \boldsymbol{x}, \boldsymbol{y}_{<t}\right)\right)$ serves as the advantage in OPD, enabling token-level credit assignment.

\subsection{Method}
\label{app:Method}

\paragraph{Mitigating Premature Convergence via Token Selection}
\label{app:Method2}
Tokens with large $h_{t}$ or $d_{t}$ are considered high-value. We combine these two criteria via the Soft-OR formulation, which approximates the logical OR operation in a differentiable manner: if either $\hat{h}_t$ or $\hat{d}_t$ is large, $s_t$ will be large, while avoiding double-counting when both are large. Given min-max normalized $\hat{h}_{t}, \hat{d}_{t} \in [0, 1]$, the selection score is:
\begin{equation}
    s_{t}= \hat{h}_{t} + \hat{d}_{t} - \hat{h}_{t} \cdot \hat{d}_{t}.
    \label{eq:st}
\end{equation}

\subsection{Experiment Details}
\label{app:Experiment Details}
The APPS dataset contains 10,000 samples in total~\citep{DBLP:conf/nips/HendrycksBKMAGB21}, spanning three difficulty levels: Introductory, Interview, and Competition. To accelerate evaluation, we uniformly sample 500 instances as our test set, with an equal number of samples drawn from each difficulty level.

\subsection{Baselines}
\label{app:Baselines}
For the configuration of naive GRPO+OPD, we follow KDRL~\citep{DBLP:journals/corr/abs-2506-02208}, which considers three variants: (1) applying OPD on all samples, (2) applying OPD only on samples where the student fails, and (3) applying OPD only on negative zero-variance prompts. For each model pair, we evaluate all three variants and select the best-performing one as the naive GRPO+OPD baseline. The OPD coefficient is kept consistent with our method.

\subsection{Implementation Details.}
\label{app:Implementation Details.}
We implement \methodname{} based on the VeRL framework~\citep{DBLP:conf/eurosys/ShengZYWZZPL025}. For mathematics, we train for 550 steps on 8 A100 GPUs; the wall-clock time for Qwen3-1.7B-Instruct $\rightarrow$ Qwen3-4B-Instruct-2507, Qwen3-4B-Instruct $\rightarrow$ Qwen3-4B-Instruct-2507, and Qwen2.5-3B-Instruct $\rightarrow$ Qwen2.5-14B-Instruct is approximately 3, 4, and 2.5 days, respectively. For code, we train for 400 steps, taking approximately 4, 5, and 3 days for the three pairs, respectively.

\paragraph{Cost Analysis}
During GRPO training, taking Qwen2.5-3B-Instruct $\rightarrow$ Qwen2.5-14B-Instruct as an example, negative zero-variance prompts account for approximately $15.6\%$ of steps initially, and this proportion decreases as the model improves. Since \methodname{} applies OPD and SFT exclusively on negative zero-variance prompts, the additional overhead over standard GRPO is minimal. On 8 A100 GPUs for 550 training steps, the extra cost amounts to approximately 12 hours, which is acceptable.

Furthermore, the SFT data is pre-generated offline and incurs no cost during training. This pre-generation process is efficient: for the 57K training set with a single rollout per prompt, it takes only about 2 hours on 2 A100 GPUs. Overall, the computational cost of \methodname{} is fully manageable.

\begin{table*}[!t]
\centering
\caption{Ablation study results on mathematical reasoning benchmarks.}
\label{tab:ablation}
\begin{tabular}{lccccc}
\toprule
\textbf{Method} & \textbf{AIME24} & \textbf{AIME25} & \textbf{MATH500} & \textbf{OLYMPIAD} & \textbf{AVG} \\
\midrule
GRPO                                        & 34.79 & 27.71 & 88.64 & 55.15 & 51.57 \\
GRPO+OPD (All)             & 35.62 & 24.79 & 86.76 & 55.55 & 50.68 \\
GRPO+OPD (Negative zero-var) & 35.21 & 26.88 & 88.27 & 55.13 & 51.37 \\
+Advantage Weighting                        & 36.88 & 28.75 & 88.74 & 56.39 & 52.69 \\
+Token Selection                            & 37.92 & 29.58 & 88.75 & 56.69 & 53.24 \\
\rowcolor{gray!10} \textbf{+SFT (Ours)}     & \textbf{42.98} & \textbf{31.87} & \textbf{89.36} & \textbf{57.36} & \textbf{55.39} \\
\bottomrule
\end{tabular}
\end{table*}

\subsection{Training Dynamics}
\label{app:Training Dynamics}
\begin{figure*}[t]
\centering
\includegraphics[width=0.96\textwidth]{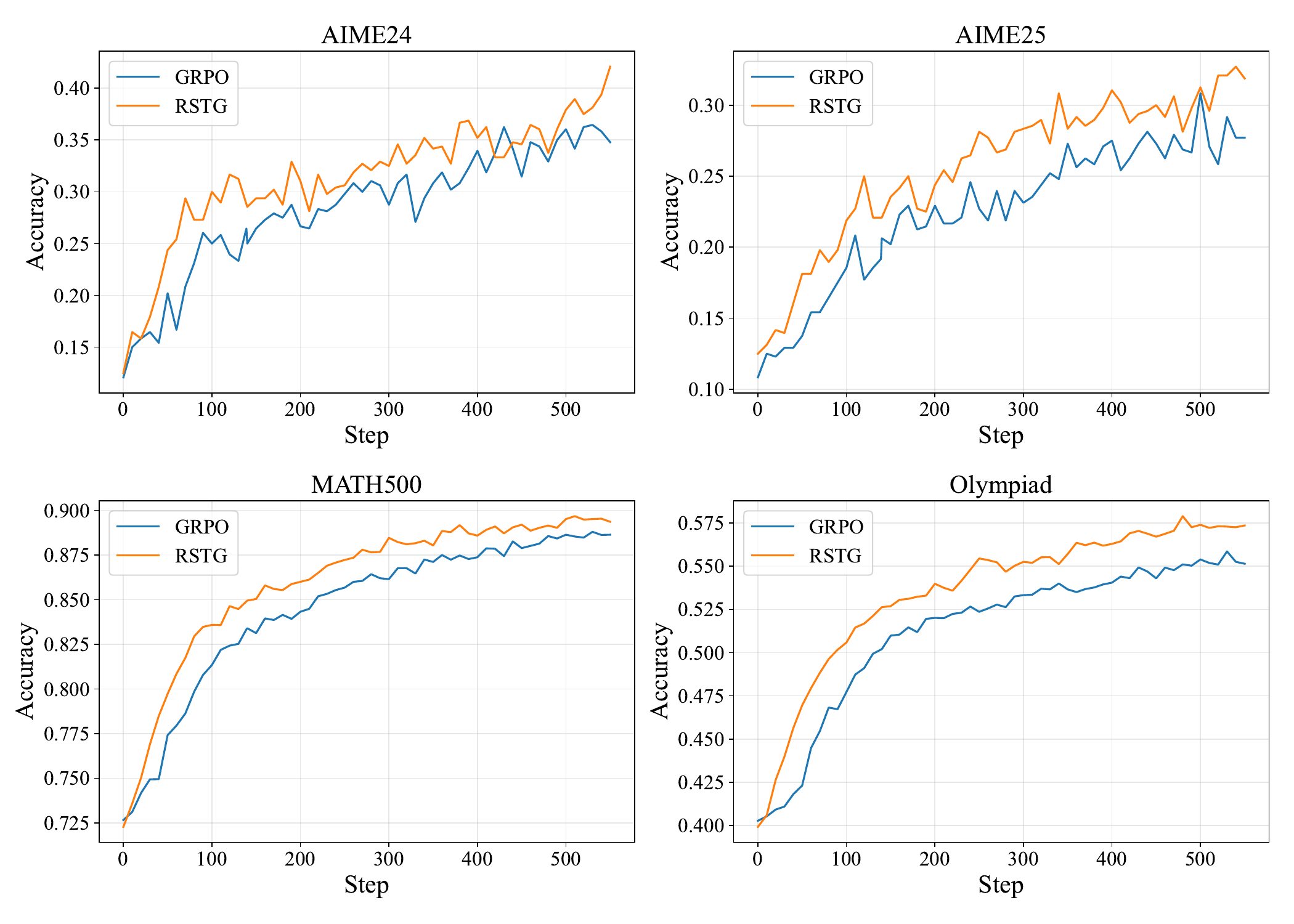} 
\caption{Performance of Qwen3-1.7B-Instruct $\rightarrow$ Qwen3-4B-Instruct-2507 on mathematical benchmarks throughout training, compared against standard GRPO.}
\label{fig:grpo_vs_rstg_4grid_1}
\end{figure*}

\begin{figure*}[t]
\centering
\includegraphics[width=0.96\textwidth]{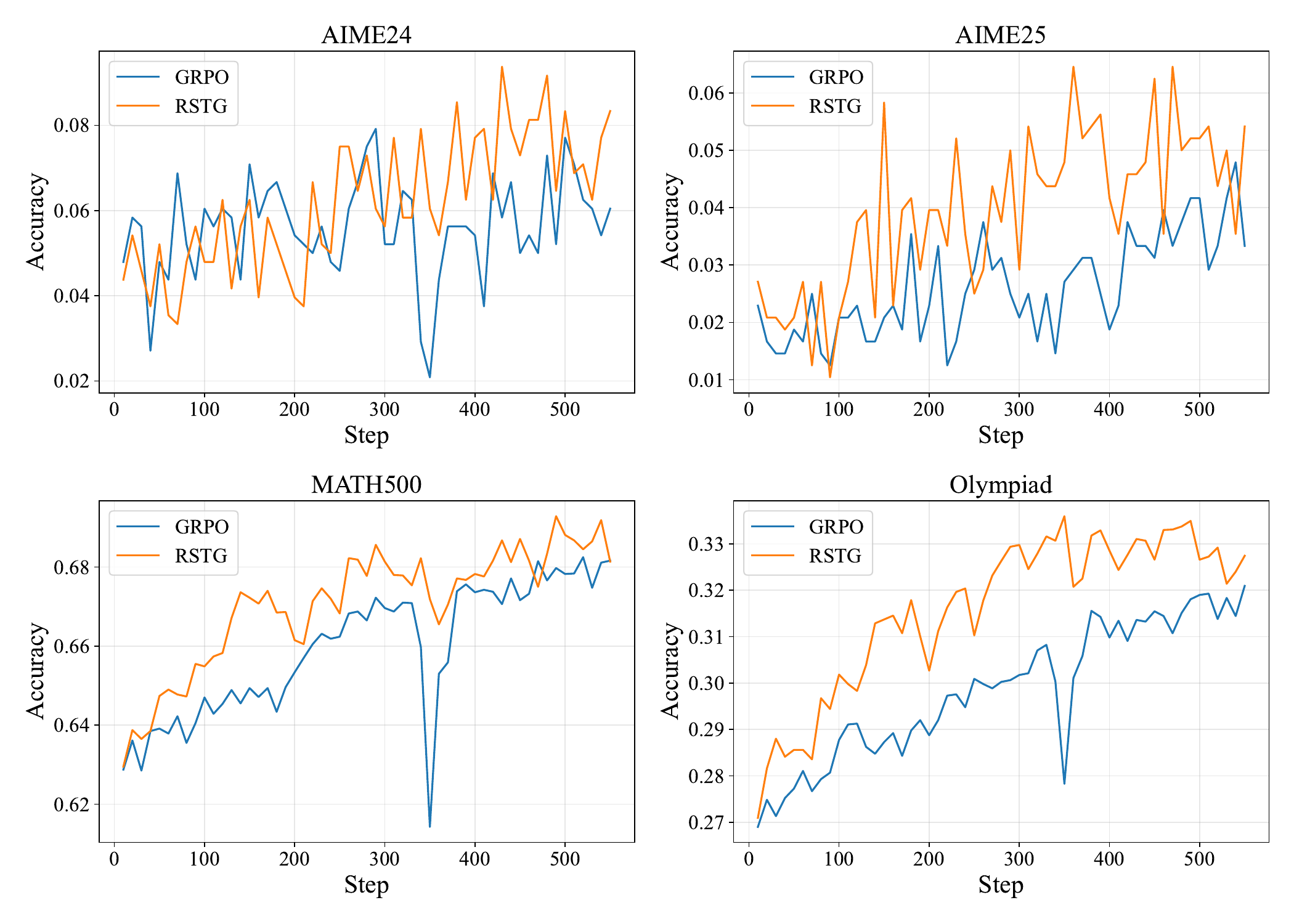} 
\caption{Performance of Qwen2.5-3B-Instruct $\rightarrow$ Qwen2.5-14B-Instruct-2507 on mathematical benchmarks throughout training, compared against standard GRPO.}
\label{fig:grpo_vs_rstg_4grid_pair3}
\end{figure*}

We select one model pair from each of the Qwen3 and Qwen2.5 families and present their performance on mathematical benchmarks throughout training in Figures~\ref{fig:grpo_vs_rstg_4grid_1} and~\ref{fig:grpo_vs_rstg_4grid_pair3}.

\subsection{Ablation Studies}
\label{app:Ablation Studies}
Taking Qwen3-1.7B-Instruct$\rightarrow$Qwen3-4B-Instruct as an example, we start from a baseline that naively combines the GRPO and OPD losses over all samples, and progressively incorporate each component of our \methodname{}. Detailed results are provided in Table~\ref{tab:ablation}, where each component is shown to contribute meaningfully to the overall performance.

\end{document}